\documentclass[final,3p,times]{elsarticle}

\usepackage{amsmath,amsfonts}
\usepackage{algorithm} 
\usepackage{algorithmic}
\usepackage{array}
\usepackage[caption=false,font=normalsize,labelfont=sf,textfont=sf]{subfig}
\usepackage{textcomp}
\usepackage{stfloats}
\usepackage{url}
\usepackage{verbatim}
\usepackage{graphicx}

\usepackage{tikz}
\usepackage{multirow} 
\usepackage{booktabs}
\usepackage[utf8]{inputenc}

\usepackage{amssymb}
\usepackage{amsmath}
\usepackage[numbers]{natbib}

\journal{}

\begin{document}

\begin{frontmatter}
	
\title{Efficient Graph Knowledge Distillation from GNNs to Kolmogorov--Arnold Networks via Self-Attention Dynamic Sampling}

\author[1]{Can Cui\fnref{orcid1}}
\ead{15583367303@163.com}

\author[1]{Zilong Fu\fnref{orcid2}}
\ead{f17610136029@163.com}

\author[1]{Penghe Huang}
\ead{hph@djtu.edu.cn}

\author[1]{Yuanyuan Li}
\ead{forkp@djtu.edu.cn}

\author[2]{Wu Deng}
\ead{dw7689@163.com}

\author[1]{Dongyan Li\corref{cor1}\fnref{orcid3}}
\ead{lidy@djtu.edu.cn}

\cortext[cor1]{Corresponding author}

\address[1]{School of Railway Intelligent Engineering, Dalian Jiaotong University, Dalian 116028, Liaoning, China}
\address[2]{College of Electronic Information and Automation, Civil Aviation University of China, Tianjin 300300, Tianjin, China}

\fntext[orcid1]{ORCID: 0009-0003-5751-8425}
\fntext[orcid2]{ORCID: 0009-0001-3087-9904}
\fntext[orcid3]{ORCID: 0000-0003-4619-8238}

\begin{abstract}
Recent success of graph neural networks (GNNs) in modeling complex graph-structured data has fueled interest in deploying them on resource-constrained edge devices. However, their substantial computational and memory demands present ongoing challenges. Knowledge distillation (KD) from GNNs to MLPs offers a lightweight alternative, but MLPs remain limited by fixed activations and the absence of neighborhood aggregation, constraining distilled performance. To tackle these intertwined limitations, we propose SA-DSD, a novel self-attention-guided dynamic sampling distillation framework. To the best of our knowledge, this is the first work to employ an enhanced Kolmogorov-Arnold Network (KAN) as the student model. We improve Fourier KAN (FR-KAN+) with learnable frequency bases, phase shifts, and optimized algorithms, substantially improving nonlinear fitting capability over MLPs while preserving low computational complexity. To explicitly compensate for the absence of neighborhood aggregation that is inherent to both MLPs and KAN-based students, SA-DSD leverages a self-attention mechanism to dynamically identify influential nodes, construct adaptive sampling probability matrices, and enforce teacher-student prediction consistency. Extensive experiments on six real world datasets demonstrate that, under inductive and most of transductive settings, SA-DSD surpasses three GNN teachers by 3.05\%–3.62\% and improves FR-KAN+ by 15.61\%. Moreover, it achieves a 16.69× parameter reduction and a 55.75\% decrease in average runtime per epoch compared to key benchmarks.
\end{abstract}



\end{frontmatter}




\section{Introduction}
\label{Introduction}
Knowledge distillation (KD) has emerged as a fundamental paradigm for transferring knowledge from complex models to compact ones, enabling efficient learning while preserving predictive performance. Beyond model compression, KD is increasingly viewed as a form of representation transfer, where the student is guided to approximate the functional behavior of a teacher under a constrained hypothesis space \cite{fang2026knowledge,razi2025comprehensive,yang2025survey}.

In the context of graph structured data, graph knowledge distillation (GKD) aims to transfer topology aware representations learned by Graph Neural Networks (GNNs) to student models that lack explicit non-Euclidean modeling capabilities \cite{tian2025knowledge}. A majority of existing GKD approaches distill knowledge into lightweight neural networks, particularly multi-layer perceptrons (MLPs), under the assumption that the student model adopts fixed pointwise activation functions and fully connected transformations.

Under this assumption, Zhang et al.~\cite{zhang2021graph} proposed GLNN, which trains an MLP using soft targets generated by a GNN to compensate for the absence of graph structure. Tan et al.~\cite{tan2023double} introduced RKD-MLP, employing a meta learning strategy to filter unreliable soft labels, albeit at the cost of further reducing the effective sample size. To alleviate this issue, Wu et al.~\cite{wu2023quantifying} proposed KRD, which quantifies the reliability of GNN knowledge via entropy based upsampling and incorporates it as a supervisory signal. Tian et al.~\cite{tian2024decoupled} introduced DGKD, which decouples the distillation loss into target class and non-target class components modulated by prediction confidence. While these methods demonstrate strong empirical performance, they inherently rely on MLP based students and thus inherit the limitations of fixed activation functions and point wise function parameterization. More critically, these approaches implicitly assume that the student model shares a functional hypothesis space similar to that of a MLP, a premise that has has received limited attention from the perspective of representation learning. In this work, we argue that the architectural form of the student model is not a secondary design choice, but a key factor that fundamentally reshapes the distillation mechanism.

\begin{figure}[tp!]
	\centering
	\includegraphics[width=0.7\linewidth]{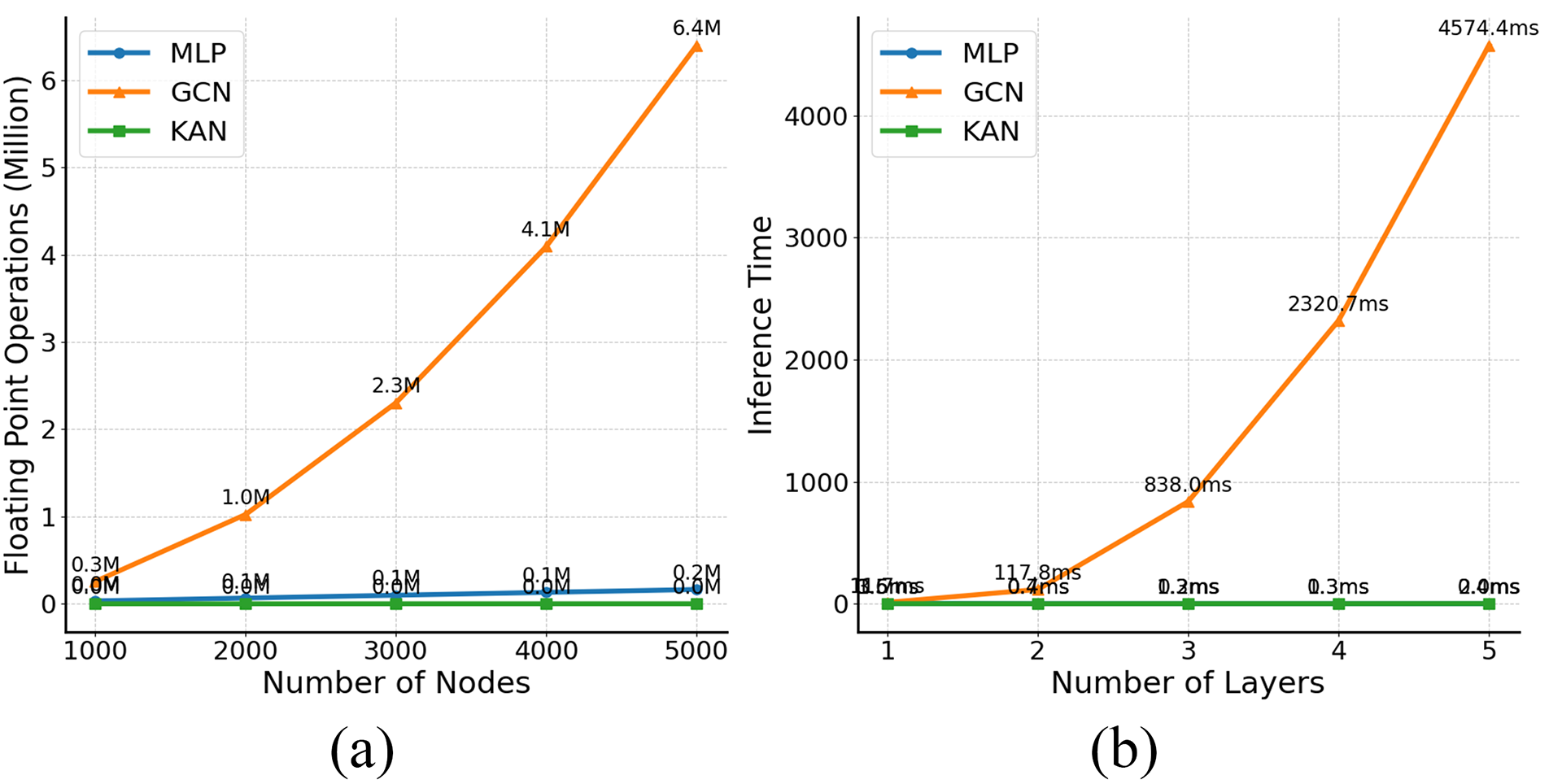}
	\caption{Visualization of (a) computational complexity comparison and (b) inference time comparison.}
	\label{fig:1}
\end{figure}

Recently, Kolmogorov–Arnold Networks (KANs) have been proposed as an alternative function approximation paradigm that replaces fixed activation functions with learnable basis functions inspired by the Kolmogorov–Arnold representation theorem. By parameterizing nonlinear mappings through edge-activated univariate functions, KANs exhibit a fundamentally different inductive bias from that of MLPs. Liu et al.~\cite{liu2024kan} first demonstrated that KANs achieve superior convergence and interpretability in lower dimension tasks, albeit with higher inference costs at scale. Subsequent studies have validated the expressive advantages of KANs across diverse domains. Guo et al. \cite{guo2025kan} applied KAN-based CQL in offline reinforcement learning, achieving performance comparable to MLPs with fewer parameters. Shi et al.~\cite{shi2025kan} proposed PointKAN for point cloud analysis, significantly outperforming PointMLP in few shot settings while reducing computational complexity. Herbozo Contreras et al.~\cite{herbozo2025kan} introduced KAN-EEG, which demonstrated strong generalization and robustness on a cross continental epilepsy dataset. These results collectively suggest that KANs offer a more expressive function parameterization than MLPs in several empirical studies. Fig.~\ref{fig:1} further compares the computational complexity and inference time of GNNs, MLPs, and KANs as model scale increases.

In terms of optimization, Li et al.~\cite{li2024kolmogorov} proposed FastKAN, approximating B-splines with Gaussian radial basis functions to achieve a 3.3x speedup while preserving accuracy. Bodner et al.~\cite{bodner2024convolutional} extended KANs to convolutional architectures, demonstrating competitive performance with CNNs and RNNs using fewer parameters. Xu et al.~\cite{xu2024fourierkan} combined Fourier transforms with KANs to propose Fourier-KAN-GCF for graph recommendation, replacing weight parameters with Fourier coefficients. These advances, particularly Fourier-based variants, motivate us to explore whether KANs’ frequency-domain parameterization could better capture the multifrequency characteristics often observed in graph representations learned by GNNs. More importantly, the role of KANs in knowledge distillation has not been systematically investigated.

When a basis function-driven representation is adopted instead of a pointwise activation structure, the behavior of knowledge distillation is fundamentally altered. This challenge becomes more pronounced in graph knowledge distillation: topology is encoded by GNNs via neighborhood aggregation, whereas KANs, similar to MLPs, do not provide explicit graph operations. Consequently, conventional distillation objectives that rely only on output matching or feature alignment are insufficient to preserve topology-relevant information; this formulation frequently leads to optimization instability and poorer generalization.

To address these challenges, we propose a Self-Attention Dynamic Sampling Distillation (SA-DSD) framework to enable effective knowledge transfer from GNNs to KAN-based student models. To the best of our knowledge, this is the first systematic study of knowledge distillation from GNNs to KAN-based students. We develop FR-KAN+ as student model, an enhanced variant of FR-KAN that integrates complex valued weights with Fourier transforms, enabling compact frequency domain representations via learnable frequency bases and phase shift parameters. While FR-KAN+ improves expressiveness and efficiency, the absence of explicit graph aggregation still limits direct distillation from GNNs. SA-DSD addresses this issue by introducing a self-attention based sampling mechanism that dynamically estimates sample importance during training. By leveraging teacher–student prediction consistency as a supervisory signal, SA-DSD guides the student to focus on the most informative samples. Importantly, prediction consistency is not used to model distributional reliability, but to approximate topology aware supervision under architectural mismatch. Extensive experiments on six benchmark datasets under both inductive and transductive settings demonstrate that the proposed method substantially improves accuracy while achieving significant model compression and inference efficiency.

The main contributions of this paper are summarized as follows:
\begin{itemize}
	\item We proposed the FR-KAN+ model, which improves the computational efficiency and frequency domain performance of the traditional FR-KAN by introducing learnable logarithmic frequency bases, complex valued weights, and phase shift parameters.
	
	\item We introduce SA-DSD, a novel distillation framework that dynamically selects informative samples via self-attention and teacher–student prediction consistency, enabling effective knowledge transfer under architectural mismatch between GNNs and KANs.

	\item Extensive experiments on six public datasets demonstrate that SA-DSD improves average accuracy by 15.61\% over FR-KAN+ and by 3.05\% to 3.62\% over three baseline GNN models. Additionally, it achieves an average compression of 16.69× and reduces average runtime per epoch by 55.75\% compared to key baselines. Ablation studies further validate the contribution of each component.
\end{itemize}

The structure of this paper is as follows: Section \ref{RELATED_WORK} reviews classic graph distillation methods, KAN networks, and FR-KAN networks; Section \ref{METHODOLOGY} introduces the FR-KAN+ model and the implementation details of the SA-DSD method; Section \ref{EXPERIMENTS} presents the experimental results in detail; Section \ref{CONCLUSION} provides the conclusion of this study.

\begin{figure*}[tp!]
	\centering
	\includegraphics[width=1\linewidth]{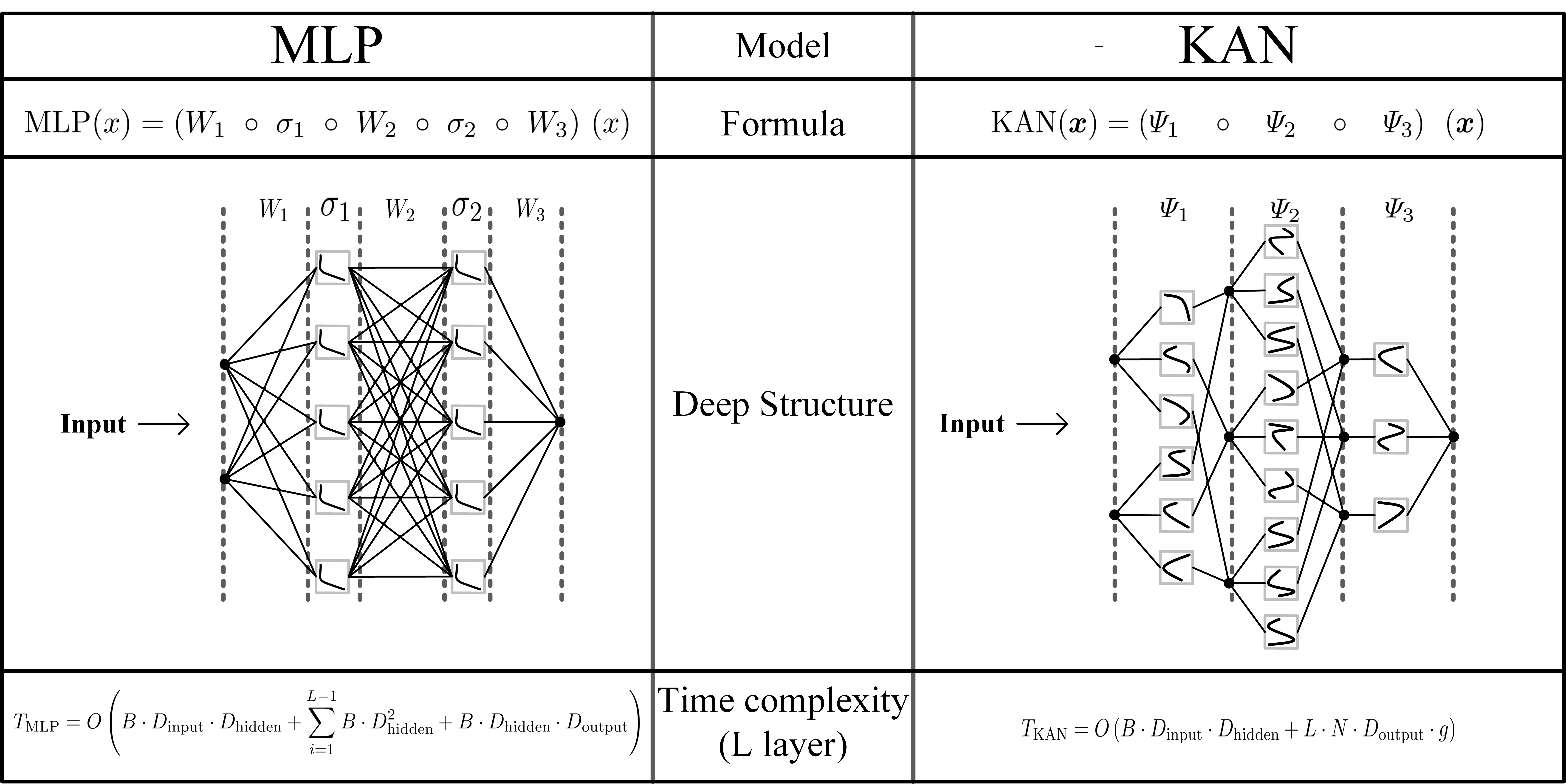}
	\caption{Architecture comparison between deep MLPs and KANs.}
	\label{fig:MLPs_vs_KANs}
\end{figure*}

\section{RELATED WORK}
\label{RELATED_WORK}
This section briefly reviews the background knowledge relevant to our research, focusing on the fundamental concepts of graphs, Graph Knowledge Distillation (GKD), Kolmogorov-Arnold Network (KAN), and Fourier KAN Network (FR-KAN).

\subsection{Basic Concepts}
A graph is typically defined as $\mathcal{G} = (\mathcal{V}, \mathcal{E})$, where $\mathcal{V}$ represents the set of nodes and $\mathcal{E}$ represents the set of edges. Let $\mathbb{N}$ denote the set of natural numbers, and $N \in \mathbb{N}$ be the number of nodes. The node feature matrix is represented as $X \in \mathbb{R}^{N \times D}$, where $N$ is the number of nodes, and $D$ is the feature dimension of each node. The adjacency matrix of the graph is represented as $A \in \mathbb{R}^{N \times N}$, where, if there is an edge between node $i$ and node $j$, $A_{i,j} = 1$; otherwise, $A_{i,j} = 0$. 

In node classification tasks, the objective is to predict the class of each node, denoted as $Y \in \mathbb{R}^{N \times K}$, where $K$ is the number of classes. Some node labels in the graph are labeled. The set of labeled nodes is denoted as $\mathcal{V}^{L}$, with corresponding feature and label matrices $X^{L}$ and $Y^{L}$, while the set of unlabeled nodes is denoted as $\mathcal{V}^{U}$, with corresponding feature and label matrices $X^{U}$ and $Y^{U}$.

\subsection{Graph Knowledge Distillation}
GKD achieves model compression and acceleration through the transfer of knowledge. The core idea of GKD involves two paradigms: GNNs-to-GNNs and GNNs-to-MLPs. In both paradigms, the teacher model is represented as $f_t(G; \theta_t)$, and the student model as $f_s(G; \theta_s)$. Knowledge transfer is achieved by minimizing the Kullback-Leibler (KL) divergence between the task loss and the soft target distribution. The standard loss function can be expressed as follows:

\begin{equation} 
	\label{eq:1}
	\mathcal{L} = \lambda\mathcal{L_{\text{task}}} + \mu D_{\text{KL}}\left( f_t \parallel f_s \right) 
\end{equation}

where $\mathcal{L_{\text{task}}}$ represents the task loss, typically the cross-entropy loss in classification tasks, $D_{\text{KL}}$ is the KL divergence, which measures the difference between the probability distributions output by the teacher model $f_t$ and the student model $f_s$, and $\lambda$ and $\mu$ are hyperparameters that control the relative contributions of the task loss and knowledge transfer loss. Existing GKD methods can be broadly categorized into GNN-to-GNN and GNN-to-MLP approaches.

A majority of existing GKD methods employ multilayer perceptron (MLP)-based students with fixed pointwise activations, which implicitly limits the student’s hypothesis space compared to more expressive architectures. While GNN-to-GNN distillation \cite{yang2020distilling,yan2020tinygnn,zhang2020reliable,feng2022freekd} preserves graph operations, GNN-to-MLP approaches dominate when targeting extreme compression.

\subsection{Kolmogorov-Arnold Networks}
The Kolmogorov-Arnold representation theorem provides a theoretical framework for approximating multivariate functions by hierarchically combining univariate functions, thereby inspiring the development of KANs. The theorem states that any continuous multivariate function $f: [0,1]^n \rightarrow \mathbb{R}$ can be represented as a sum of multiple univariate functions, as demonstrated in \cite{liu2024kan}:
\begin{equation} 
	\label{eq:2}
	f(\boldsymbol{x}) = \sum_{q=1}^{2n+1} \Phi_q \left( \sum_{p=1}^n \Psi_{q,p}(x_p) \right) 
\end{equation}
where ${\Phi_q}$ and ${\Psi_{q,p}}$ represent the outer and inner function groups, respectively, corresponding to the nonlinear transformation of the input dimension $x_p$. This decomposition overcomes the limitations of traditional MLPs with fixed activation functions, enabling high-dimensional mappings through adaptive combinations of functions. The KAN model is parameterized through trainable basis functions, utilizing a linear combination of B-spline basis functions and the SiLU activation function:

\begin{equation} 
	\label{eq:3}
	\Psi_{q,p}(x) = \omega_{q,p} \text{SiLU}(x) + \sum_{k=1}^K c_{q,p,k} B_k(x)
\end{equation}

where $B_k(\cdot)$ represents the B-spline basis functions, ${c_{q,p,k}}$ are the learnable spline coefficients, and $\omega_{q,p}$ controls the linear activation component. This approach supports gradient-based optimization and retains its general approximation capabilities during training. The deep KAN architecture extracts features through hierarchical composition, as expressed by the following formula \cite{liu2024kan}:
\begin{equation}
	\label{eq:4}
	\text{KAN}(\boldsymbol{x}) = (\varPsi_{L} \circ \varPsi_{L-1} \circ \cdots \circ \varPsi_{1})(\boldsymbol{x})
\end{equation}

where $L$ is network depth, and each layer $\varPsi_l$ learns linear and nonlinear transformations adaptively through parameterized univariate functions.  

Fig.\ref{fig:MLPs_vs_KANs} provides a visual representation of the structural differences between KAN and MLP when both the number of layers and the grid size are set to 3 \cite{liu2024kan}, along with a comparison of their time complexities under a multilayer network with \( L \) layers. Since KAN utilizes edge activation, it can effectively reduce redundant calculations and enhance the model's ability to capture complex patterns when compared to the traditional MLP structure. This design not only improves computational efficiency but also boosts the model's performance in handling high-dimensional data. From a distillation perspective, KANs differ from MLPs in that their learnable basis functions enable function level adaptation during knowledge transfer, rather than relying solely on fixed point wise activations.

\subsection{FR-KAN Model}
The traditional KAN model is more difficult to train than MLPs due to its reliance on spline functions for nonlinear approximation. This requires multiple condition checks and iterative steps, thereby increasing training complexity and computational costs. Additionally, the grid update mechanism may cause instability with uneven data distributions.

Since the core idea of KAN is to approximate functions by summing nonlinear components, replacing spline functions with Fourier coefficients preserves complex relationships while enabling more efficient function transformations. Several recent works have replaced B-splines with Fourier bases to improve training efficiency, resulting in formulations \cite{xu2024fourierkan,ai2024grokformer,imran2024fourierkan}, resulting in the following formulation:
\begin{equation}
	\label{eq:5}
	\varPsi_F(x) = \displaystyle\sum_{i=1}^{D} \displaystyle\sum_{k=1}^{g} (a_{ik} \cdotp cos(kx_{i} ) + b_{ik} \cdotp sin(kx_{i}) )
\end{equation}

where $a_{ik}$ and $b_{ik}$ are the $i$-th trainable Fourier coefficients, $g$ is the grid size, which determines the frequency terms used in the Fourier series expansion, $D$ is the input feature dimension, and $x_i$ is the $i$-th feature dimension.


\section{METHODOLOGY}
\label{METHODOLOGY}
\begin{figure*}[tp!]
	\centering
	\includegraphics[width=0.9\linewidth]{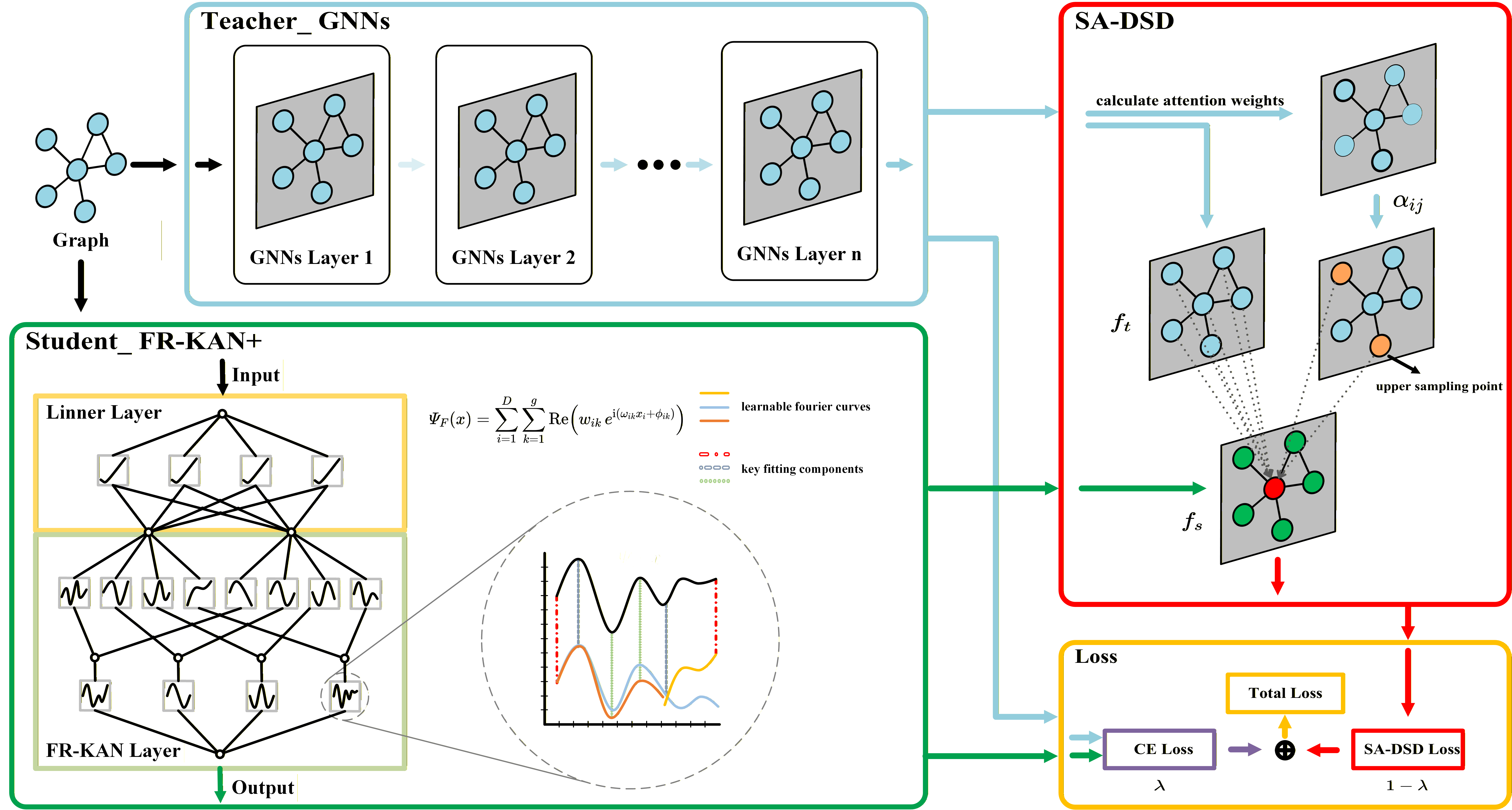}
	\caption{Overall framework diagram of SA-DSD.}
	\label{fig:framework}
\end{figure*}

In this section, we provide a detailed description of the improved FR-KAN+ model and its application in the SA-DSD distillation method. The overall framework of the method is presented in Fig.\ref{fig:framework}. Specifically, the green-boxed area illustrates the architecture of the FR-KAN+ model, while the red-boxed area highlights the schematic of the SA-DSD distillation process.

\subsection{FR-KAN+ Model}
While the traditional Fourier KAN model improves interpretability and execution efficiency, it still encounters challenges in handling complex nonlinear relationships, high-dimensional data, and training stability. To address these limitations, we have enhanced the Fourier KAN model.

First, the frequency of the Fourier series is made dynamic, generated through learnable frequency basis parameters. A dynamic frequency basis $\omega_k$ is introduced, and the frequency is adjusted using a logarithmic scale. Specifically, the dynamic frequency basis is mapped to actual frequency values via a learnable logarithmic frequency basis $\log \omega_{k}$, with its distribution range and scaling adjusted to suit different data distributions.

Next, the FR-KAN+ model combines $a_{ik}$ and $b_{ik}$ into complex weights $w_{ik} = a_{ik} + i b_{ik}$ and simplifies the expression of the Fourier basis function $e^{i k x_i}$ using Euler's formula. An additional learnable phase shift $\phi_{ik}$ is introduced, allowing the phase of the Fourier basis function for each input feature to be adjusted flexibly, thus enhancing the model's expressive capability. This results in the following equation \eqref{eq:6}:

\begin{gather}
	\label{eq:6}
	\varPsi_F(x)=\sum_{i=1}^{D}\sum_{k=1}^{g}\operatorname{Re}\!\left(w_{ik}\,e^{\mathrm{i}\left(\omega_{ik}x_i+\phi_{ik}\right)}\right),\\
	\omega_{ik}=\exp(\tilde{\omega}_{ik}) \notag
\end{gather}

where $w_{ik}=a_{ik}+\mathrm{i}b_{ik}\in\mathbb{C}$ denotes the learnable complex weight,
$\phi_{ik}$ is a learnable phase shift, and $\omega_{ik}$ is the learnable dynamic frequency.
We parameterize $\omega_{ik}$ in the log-frequency space via $\omega_{ik}=\exp(\tilde{\omega}_{ik})$
to ensure $\omega_{ik}>0$ and improve training stability.

Finally, tensor contraction between complex weights and Fourier basis functions is efficiently performed using the einsum operation, replacing the complex checks and iterative steps required by traditional spline functions. By introducing periodic variation and dynamically capturing input features, the FR-KAN+ model improves computational efficiency while enhancing its ability to model complex nonlinear relationships.

\subsection{SA-DSD Distillation Method}
We employ a $Query$-$Key$ mechanism to compute node-wise attention weights, which serve as a teacher-guided prior for reweighting distillation signals.
Different from using the raw node features directly, we take the teacher logits as the attention input, i.e., $X_t = f_t(G, X) \in \mathbb{R}^{N \times C}$, where $N$ is the number of nodes and $C$ is the number of classes.
We first apply two linear transformations to obtain the $Query$ and $Key$ representations:
\begin{equation}
	\label{eq:7}
	[Q, K] = X_t [W_Q, W_K] + [b_Q, b_K]
\end{equation}
where $W_Q, W_K \in \mathbb{R}^{C \times H}$ are learnable projection matrices, $b_Q, b_K \in \mathbb{R}^{H}$ are bias vectors, and $H$ is the mapped feature dimension.

The attention scores are computed by scaled dot product attention and normalized by softmax:
\begin{equation}
	\label{eq:8}
	\alpha_{ij} = softmax_j(\frac{Q_i^TK_j}{\sqrt{H} } )
\end{equation}
where $\alpha_{ij}$ denotes how much node $i$ attends to node $j$.
To obtain node level importance, we aggregate the attention received by each node via a column wise summation:
\begin{equation}
	\label{eq:9}
	w_j = \sum_{i=1}^{N} \alpha_{ij}, \quad \tilde{w}_j = \frac{w_j}{\max_k w_k + \epsilon}
\end{equation}
where $\tilde{w}_j \in [0,1]$ is the normalized attention weight. We set a small $\epsilon$ to avoid division by zero.
Max-normalization stabilizes the scale of $\tilde{w}$, while the subsequent step forms proper probability distributions through group wise normalization.

Let $X_s = f_s(X) \in \mathbb{R}^{N \times C}$ denote the student logits, and let $x_i^t$ and $x_i^s$ be the $i$-th row of $X_t$ and $X_s$, respectively. Based on the prediction consistency between teacher and student, we further construct a dynamic node weighting distribution for distillation.
Let $c_i = \mathbb{I}\big[\arg\max x_i^t = \arg\max x_i^s\big]$ be the consistency indicator.
We define the node probability $p_i$ as a mixture of two normalized distributions:
\begin{equation}
	\label{eq:10}
	p_i =
	\begin{cases}
		(1-\gamma)\dfrac{\tilde{w}_i}{\sum_{k:c_k=1}\tilde{w}_k}, & c_i=1,\\[8pt]
		\gamma\dfrac{\tilde{w}_i}{\sum_{k:c_k=0}\tilde{w}_k}, & c_i=0,
	\end{cases}
\end{equation}
where $\gamma \in (0,1)$ controls the relative emphasis between consistent and inconsistent nodes in the distillation process.
In practice, if either group is empty, we skip the corresponding normalization and fall back to the attention prior to ensure numerical stability.
Unless otherwise specified, the normalizations in Eq.\eqref{eq:10} are computed over the training nodes.
This design combines a teacher-guided attention prior $\tilde{w}$ with a consistency aware reweighting strategy, enabling the student to prioritize more informative node-wise distillation signals.

\subsection{Design of the Loss Function}
To achieve knowledge distillation from GNN to FR-KAN+, the total loss consists of two parts. The first part is the cross-entropy loss between the student model and the labels, defined as:
\begin{equation}
	\label{eq:11}
	\mathcal{L}_{CE} = \mathbin{-}\frac{1}{|\mathcal{T}|}\sum_{i \in \mathcal{T}} \sum_{c=1}^{C} y_{ic} \log(\hat{y}_{ic})
\end{equation}
where $y_{ic}$ is the ground-truth one-hot label, $\hat{y}_{ic}$ is the predicted probability of the student model for class $c$, and $\mathcal{T}$ denotes the set of training nodes.

We compute the distribution difference between the teacher and student models on training nodes, weighted by the node probabilities $p_i$ in Eq.\eqref{eq:10}:
\begin{equation}
	\label{eq:12}
	\mathcal{L}_{SA-DSD} = \sum_{i \in \mathcal{T}} p_i \cdot D_{KL}\Big( \sigma ( \frac{x_i^t}{\tau} ) \parallel \sigma ( \frac{x_i^s}{\tau} ) \Big)
\end{equation}
where $\tau$ is the distillation temperature that smooths the output distribution, and $\sigma(\cdot)$ is the softmax operator.
Following common KD practice, we use temperature $\tau$ in the softmax. We do not apply the additional $\tau^2$ gradient scaling factor and absorb it into the loss balancing coefficient.

Finally, the overall objective is:
\begin{equation}
	\label{eq:13}
	\mathcal{L}_{total} = \lambda \mathcal{L}_{CE} + (1 - \lambda)\mathcal{L}_{SA-DSD}
\end{equation}
where $\lambda$ balances the original task loss and the distillation loss.

By reweighting node-wise distillation signals using attention-guided and consistency aware probabilities, we propose a non-typical decoupled distillation strategy. In our implementation, $p_i$ is used as a weighting distribution rather than requiring explicit stochastic node sampling, so Eq.\eqref{eq:12} performs full set weighted distillation over $\mathcal{T}$.

This strategy selectively emphasizes and down-weights node-wise distillation signals via the attention mechanism, reducing the reliance on explicit structural coupling while still leveraging teacher predictions to guide the student. The pseudocode for SA-DSD is summarized in Algorithm \ref{algorithmic_1}.

\begin{algorithm}[htbp]
	\caption{SA-DSD Distillation Process}
	\label{algorithmic_1}
	\begin{algorithmic}[1]
		\REQUIRE Input feature matrix $X$, true labels $y$, teacher model $f_t$, student model $f_s$, distillation temperature $\tau$, balancing factor $\lambda$, mixture factor $\gamma$.
		\ENSURE Total loss $\mathcal{L}_{total}$ used to optimize the student model.
		
		\STATE Obtain teacher logits $X_t = f_t(G, X)$ and compute $Q, K$ using Eq.\eqref{eq:7}.
		\STATE Compute attention weights $\alpha_{ij}$ using Eq.\eqref{eq:8}, and aggregate them to node importance $\tilde{w}$ using Eq.\eqref{eq:9}.
		\STATE Initialize $p_i \leftarrow 1$ for all nodes.
		
		\FOR{$epoch \in \{1, 2, \ldots, n\}$}
		\STATE Obtain logits from teacher and student: $X_t = f_t(G,X)$ and $X_s = f_s(X)$.
		\STATE Compute prediction consistency $c_i = \mathbb{I}\big[\arg\max x_i^t = \arg\max x_i^s\big]$.
		\STATE Update node weights $p_i$ using Eq.\eqref{eq:10} (with a stable fallback when a group is empty).
		\STATE Compute $\mathcal{L}_{CE}$ using Eq.\eqref{eq:11}.
		\STATE Compute $\mathcal{L}_{SA-DSD}$ using Eq.\eqref{eq:12}.
		\STATE Combine losses into $\mathcal{L}_{total}$ using Eq.\eqref{eq:13}.
		\ENDFOR
		\RETURN $\mathcal{L}_{total}$
	\end{algorithmic}
\end{algorithm}

\section{EXPERIMENTS}
\label{EXPERIMENTS}
This section first presents the datasets used in the experiments, followed by a detailed description of the baseline methods and experimental setup. We then provide a comparison of SA-DSD with the main baseline methods, highlighting its advantages in computational efficiency and performance. Additionally, we compare SA-DSD with state of the art graph knowledge distillation methods. Finally, we validate the effectiveness of each SA-DSD component through ablation experiments and visualization analysis.

\subsection{Experiment settings}

\begin{table}[htbp]
	\scriptsize
	\centering
	\caption{Details of datasets used in the experiments.}
	\label{tab:datasets}
	\begin{tabular}{lcccccccc}
		\hline\hline
		\textbf{Data Sets} & \textbf{\#Nodes} & \textbf{\#Edges} & \textbf{\#Class} & \textbf{\#Features}\\ 
		\hline
		\textbf{Cora}              & 2,708            & 5,278            & 7               & 1,433\\ 
		\textbf{Citeseer}           & 3,327            & 4,614            & 6               & 3,703\\ 
		\textbf{PubMed}             & 19,717           & 44,324           & 3               & 500\\ 
		\textbf{Photo}              & 7,650            & 119,081          & 8               & 745\\ 
		\textbf{CS}                 & 18,333           & 81,894           & 15               & 6,805\\ 
		\textbf{Physics}            & 34,493           & 247,962          & 5               & 8,415\\ 
		\hline\hline
	\end{tabular}
\end{table}

We conduct experiments on six real world benchmark datasets, detailed in Table \ref{tab:datasets}, categorized into citation networks and Amazon/Coauthor networks:

\begin{itemize} 
	\item Cora \cite{sen2008collective}, Citeseer \cite{giles1998citeseer}, and PubMed \cite{sen2008collective} are widely used citation network datasets for node
	classification tasks . These datasets contain
	academic papers and their citation relationships, where nodes represent papers and edges
	represent citation links between papers. Each paper is labeled with a research topic category.
	
	\item Amazon-Photo, Coauthor-CS, and Coauthor-Physics are benchmark graph datasets
	introduced by Shchur et al. \cite{shchur2018pitfalls}. These datasets are constructed from
	product co-purchase networks and academic co-authorship networks, respectively. Nodes
	represent products or authors, and edges represent co-purchase or collaboration relationships.
	They are commonly used for evaluating node classification methods.
	
\end{itemize}

Dataset splitting and usage follow the strategies outlined in previous works \cite{zhang2021graph, yang2021extract, wu2023quantifying} to ensure fairness and accuracy in the experiments.

\subsection{Baselines and Training Details}
To validate the model compatibility of the SA-DSD method, we selected three GNN models as teacher models: GCN \cite{kipf2016semi}, GraphSAGE \cite{hamilton2017inductive}, and GAT \cite{velivckovic2017graph}, applying them to the knowledge distillation framework. The student models selected were the FR-KAN+ model and the MLP model. This paper focuses on the distillation design from GNNs to KANs; hence, we selected the advanced GNN-to-MLP distillation method, KRD, as the primary benchmark. Additionally, we compare SA-DSD with representative GNN-to-GNN distillation baselines, including LSP \cite{yang2020distilling}, TinyGNN \cite{yan2020tinygnn}, RDD \cite{zhang2020reliable}, and FreeKD \cite{feng2022freekd}. 
We also compare SA-DSD with GNN-to-MLP baselines, including CPF \cite{yang2021extract}, RKD-MLP \cite{tan2023double}, and FF-G2M \cite{wu2023extracting}, to evaluate the feasibility and efficiency of GNN-to-KAN distillation.

To comprehensively evaluate our method, the experimental design includes both transductive and inductive settings. In the transductive setting, the model is trained based on the feature matrix \( X \) and the labeled node label matrix \( Y^{L} \) , and then infers the labels \( Y^{U} \) for the unlabeled nodes. In the inductive setting, the training and test sets are completely distinct. After the model is trained using \( X^{L}, X^{U}_{\text{obs}} \), and \( Y^{L} \), it predicts the labels \( Y^{U}_{\text{ind}} \) of the unseen unlabeled nodes. 

The experiments are conducted on the PyTorch and Deep Graph Library (DGL) platforms, with all model parameters automatically optimized using Optuna to determine the best configuration. For full reproducibility, the detailed hyperparameter configurations of SA-DSD are provided in Appendix~\ref{app:hyperparams}. To ensure repeatability and fairness, we run each method five times with five different fixed random seeds, and report the mean ± std. All models are trained using the Adam optimizer, and experiments are conducted on a single RTX 2080 Ti GPU.

\begin{table*}[htbp!]
	\scriptsize
	\centering
	\label{tab:total_table}
	\caption{Classification Accuracy ± std (\%) for Learning Three Different Teacher Models of GNNs in Transduction and Induction Modes.}
	\setlength{\tabcolsep}{4pt} 
	
	\begin{tabular}{cc|ccccc|ccccc}
		\hline	\hline
		\multirow{2}{*}{\textbf{Datasets}} & \multirow{2}{*}{\textbf{Model}} & \multicolumn{5}{c|}{\textbf{Transductive}}                                          & \multicolumn{5}{c}{\textbf{Inductive}}                           \\ \cline{3-12} 
		
		&                                 &  \textbf{Self} & \textbf{KRD} & \textbf{SA-DSD}     & \textbf{$\Delta$self} & \textbf{$\Delta$KRD} & \textbf{Self} & \textbf{KRD} & \textbf{SA-DSD}     & \textbf{$\Delta$self} & \textbf{$\Delta$KRD} \\  \hline
		\multirow{4}{*}{\textbf{cora}}     & \textbf{FR-KAN+}                & 59.82±0.26    & -            & -                   & -              & -             & 60.46±0.49    & -            & -                   & -              & -             \\
		& \textbf{GCN}                    & 81.42±0.99    & 84.1±0.87    & \textbf{85.22±0.66} & 4.67\%↑        & 1.33\%↑       & 79.62±0.41    & 74.4±0.23    & \textbf{74.93±0.71} & 5.89\%↓        & 0.71\%↑       \\
		& \textbf{SAGE}                   & 81.44±0.58    & 84.56±1.23   & \textbf{85.14±0.96} & 4.54\%↑        & 0.68\%↑       & 80.96±0.21    & 72.2±0.64    & \textbf{74.34±1.07} & 8.18\%↓        & 2.96\%↑       \\
		& \textbf{GAT}                    & 80.72±0.69    & 83.74±0.55   & \textbf{84.88±0.59} & 5.15\%↑        & 1.36\%↑       & 81.26±0.36    & 72.52±0.84   & \textbf{73.92±0.68} & 9.03\%↓        & 1.93\%↑       \\ \hline
		\multirow{4}{*}{\textbf{citeseer}} & \textbf{FR-KAN+}                & 60.28±0.58    & -            & -                   & -              & -             & 60.98±0.51    & -            & -                   & -              & -             \\
		& \textbf{GCN}                    & 71.44±0.32    & 75.26±0.37   & \textbf{75.33±1.18} & 5.45\%↑        & 0.09\%↑       & 71.8±0.32     & 71.88±0.6    & \textbf{72.42±0.41} & 0.86\%↑        & 0.75\%↑       \\
		& \textbf{SAGE}                   & 70.7±0.14     & 74.42±0.55   & \textbf{74.75±1.82} & 5.73\%↑        & 0.44\%↑       & 70.7±0.39     & 71.72±0.28   & \textbf{72.3±1.46}  & 2.26\%↑        & 0.81\%↑       \\
		& \textbf{GAT}                    & 72.14±0.38    & 72.52±1.68   & \textbf{73.81±0.72} & 2.32\%↑        & 1.78\%↑       & 69.84±0.63    & 70.3±1.2     & \textbf{70.8±1.5}   & 1.37\%↑        & 0.71\%↑       \\ \hline
		\multirow{4}{*}{\textbf{pubmed}}   & \textbf{FR-KAN+}                & 74.82±0.47    & -            & -                   & -              & -             & 74.84±0.19    & -            & -                   & -              & -             \\
		& \textbf{GCN}                    & 77.72±0.41    & 82.14±0.52   & \textbf{82.81±0.47} & 6.55\%↑        & 0.82\%↑       & 77.86±0.13    & 81.68±0.25   & \textbf{81.68±0.19} & 4.91\%↑        & $\approx$0\%          \\
		& \textbf{SAGE}                   & 76.8±0.24     & 81.28±0.40   & \textbf{82.4±0.48}  & 7.29\%↑        & 1.38\%↑       & 77.7±0.46     & 82.12±0.48   & \textbf{82.59±0.63} & 6.29\%↑        & 0.57\%↑       \\
		& \textbf{GAT}                    & 77.32±0.66    & 82.2±0.48    & \textbf{82.71±0.65} & 6.97\%↑        & 0.62\%↑       & 77.04±0.19    & 81.7±0.57    & \textbf{82.16±0.43} & 6.64\%↑        & 0.56\%↑       \\ \hline
		\multirow{4}{*}{\textbf{photo}}    & \textbf{FR-KAN+}                & 77.88±3.54    & -            & -                   & -              & -             & 77.26±4.88    & -            & \textbf{-}          & -              & -             \\
		& \textbf{GCN}                    & 89.3±0.88     & 92.22±2.14   & \textbf{93.48±1.45} & 4.68\%↑        & 1.37\%↑       & 89.74±0.61    & 91.13±2.52   & \textbf{92.09±1.65} & 2.62\%↑        & 1.05\%↑       \\
		& \textbf{SAGE}                   & 88.92±0.37    & 92.24±2.08   & \textbf{93.07±1.71} & 4.67\%↑        & 0.90\%↑       & 89.16±0.36    & 90.96±1.56   & \textbf{91.29±1.98} & 2.39\%↑        & 0.36\%↑       \\
		& \textbf{GAT}                    & 90.84±0.20    & 92.23±1.35   & \textbf{93.39±1.79} & 2.81\%↑        & 1.26\%↑       & 89.45±0.25    & 91.41±1.52   & \textbf{92.14±1.39} & 3.01\%↑        & 0.80\%↑       \\ \hline
		\multirow{4}{*}{\textbf{cs}}       & \textbf{FR-KAN+}                & 89.77±0.38    & -            & \textbf{-}          & -              & -             & 90.49±1.20    & -            & -                   & -              & -             \\
		& \textbf{GCN}                    & 90.76±1.34    & 93.86±0.38   & \textbf{94.11±0.46} & 3.69\%↑        & 0.27\%↑       & 90.25±1.67    & 93.09±0.49   & \textbf{93.66±0.4}  & 3.78\%↑        & 0.61\%↑       \\
		& \textbf{SAGE}                   & 89.97±1.59    & 93.81±0.11   & \textbf{93.91±0.58} & 4.38\%↑        & 0.11\%↑       & 89.24±0.53    & 93.00±0.77   & \textbf{94.14±0.36} & 5.49\%↑        & 1.23\%↑       \\
		& \textbf{GAT}                    & 89.21±1.30    & 94.52±0.1    & \textbf{94.5±0.52}  & 5.93\%↑        & 0.02\%↓       & 90.88±1.33    & 93.07±0.31   & \textbf{94±0.57}    & 3.43\%↑        & 1.00\%↑       \\ \hline
		\multirow{4}{*}{\textbf{physics}}  & \textbf{FR-KAN+}                & 90.33±0.64    & -            & \textbf{-}          & -              & -             & 90.47±0.80    & -            & -                   & -              & -             \\
		& \textbf{GCN}                    & 92.44±0.26    & 94.70±0.37   & \textbf{95.31±0.29} & 3.10\%↑        & 0.64\%↑       & 92.35±0.49    & 94.46±0.52   & \textbf{94.37±0.50} & 2.19\%↑        & 0.09\%↓       \\
		& \textbf{SAGE}                   & 92.05±0.72    & 94.4±0.47    & \textbf{95.07±1.10} & 3.28\%↑        & 0.71\%↑       & 91.76±1.25    & 93.34±0.72   & \textbf{94.34±0.52} & 2.81\%↑        & 1.07\%↑       \\
		& \textbf{GAT}                    & 92.43±0.45    & 94.39±0.44   & \textbf{94.56±0.57} & 2.30\%↑        & 0.18\%↑       & 91.27±0.71    & 93.84±0.58   & \textbf{94.41±0.49} & 3.44\%↑        & 0.61\%↑       \\ \hline	\hline
	\end{tabular}
\end{table*}

\begin{figure}[hbtp!]
	\centering
	\includegraphics[width=0.7\linewidth]{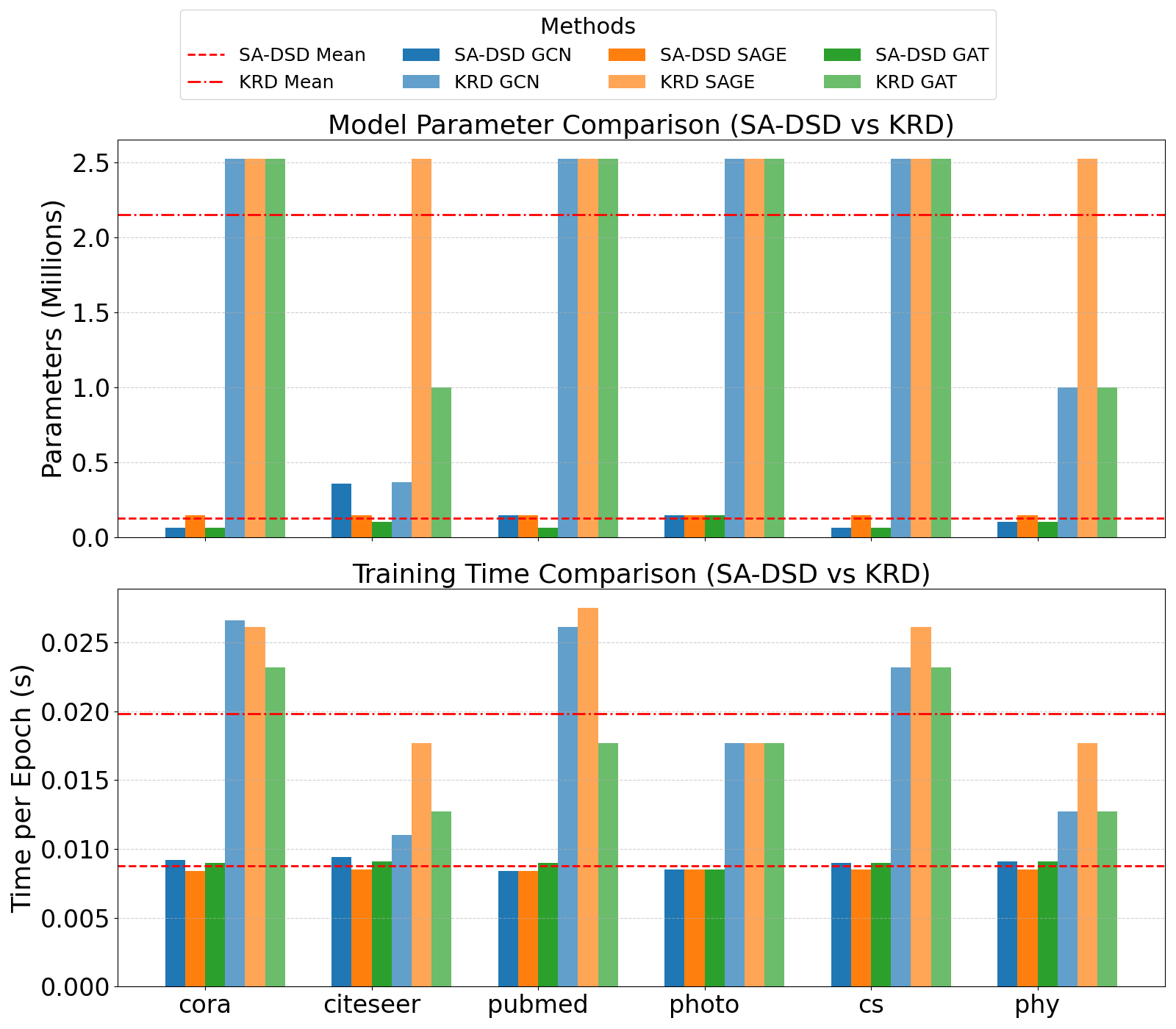}
	\caption{SA-DSD vs. KRD in terms of number of parameters and runtime.}
	\label{fig:param_and_time}
\end{figure}

\subsection{Comparison of Classification Performance}
To validate the classification performance of SA-DSD, we compared it with the baseline method KRD across six datasets, employing three different teacher GNN models to test its compatibility. The experimental results, presented in Table \ref{tab:total_table}, indicate that SA-DSD improves accuracy by 2.3\% to 7.29\% over traditional GNN models in the transductive setting, and by 0.86\% to 6.64\% in the inductive setting. These results demonstrate that SA-DSD effectively transfers knowledge from the teacher model to the student model, thereby improving classification performance in both modes. Compared to the KRD method, SA-DSD improves accuracy by 0.09\% to 1.78\% in the transductive setting and by 0.36\% to 2.96\% in the inductive setting across all datasets.

From a broader perspective, SA-DSD performs better in the transductive setting than in the inductive setting. This is because inductive learning requires the model to learn from only a subset of nodes in the training set and infer on unseen nodes, increasing the difficulty of the classification task. It is important to note that the Cora dataset has fewer nodes and contains seven classes with an imbalanced class distribution. In the inductive mode, the connectivity information of test set nodes is more limited. Both the SA-DSD and KRD distillation mechanisms fail to effectively learn enough generalized features, thus their performance is lower than that of GNNs models, which can learn rich local graph structural information through neighborhood aggregation.

The experiment also provides a detailed comparison of the parameter requirements and runtime between SA-DSD and KRD under different teacher models. As shown in Fig.\ref{fig:param_and_time}, SA-DSD significantly outperforms KRD in both parameter size and training time. Specifically, SA-DSD reduces the average number of parameters by 16.69 times compared to KRD, with a maximum compression ratio of 37.97x. In terms of time, the average runtime per epoch for SA-DSD is reduced by 55.75\% compared to KRD, with a maximum reduction of 69.45\%. This significant reduction in the size of the parameters can significantly reduce the memory storage requirements, demonstrated its potential for edge deployment.	

\subsection{Comparison with Representative Baselines}
To evaluate the performance of SA-DSD compared to other graph knowledge distillation methods, we conducted multiple experiments, including SA-DSD, FR-KAN+, GCN, and KRD. Results of LSP, TinyGNN, RDD, FreeKD, GLNN, CPF, RKD-MLP, FF-G2M are reported from \cite{wu2023quantifying} under the same transductive setting. The results in Table \ref{tab:all_compare} can be directly compared with the results in the literature \cite{wu2023quantifying}.

As shown in Table \ref{tab:all_compare}, SA-DSD outperforms other methods on all datasets, except for the CS dataset, where its performance is slightly lower than that of the RDD method. Moreover, SA-DSD achieves the highest average rank, indicating that it effectively enhances the performance of FR-KAN+ while transferring graph structural information. This superior balance of performance and efficiency makes SA-DSD highly promising for deployment on edge devices with limited computational resources.

\begin{table*}[htbp!]
	\scriptsize
	\centering
	\label{tab:all_compare}
	\caption{Classification Accuracy ± std (\%) of SA-DSD vs. other KD baseline methods.}
	\setlength{\tabcolsep}{4pt} 
	\begin{tabular}{cc|cccccc|c}
		\hline\hline
		\multicolumn{1}{c}{\textbf{Category}} & \textbf{Methods} & \textbf{cora}       & \textbf{citeseer}   & \textbf{pubmed}     & \textbf{photo}      & \textbf{cs}         & \textbf{physics}    & \textbf{Avg. Rank} \\ \hline
		\multirow{2}{*}{\textbf{Vanilla}}      & \textbf{FR-KAN+} & 59.82±0.26          & 60.28±0.58          & 74.82±0.47          & 77.88±3.54          & 89.77±0.38          & 90.33±0.64          & 13                 \\ \cline{3-9} 
		& \textbf{GCN}     & 81.42±0.99          & 71.44±0.32          & 77.72±0.41          & 89.3±0.88           & 90.76±1.34          & 92.44±0.26          & 12                 \\ \hline
		\multirow{5}{*}{\textbf{GNN-to-GNN}}   & \textbf{LSP}     & 82.70±0.43          & 72.68±0.62          & 80.86±0.50          & 91.74±1.42          & 92.56±0.45          & 92.85±0.46          & 9                  \\ \cline{3-9} 
		& \textbf{GNN-SD}  & 82.54±0.36          & 72.34±0.55          & 80.52±0.37          & 91.83±1.58          & 91.92±0.51          & 93.22±0.66          & 9.5                \\ \cline{3-9} 
		& \textbf{TinyGNN} & 83.10±0.53          & 73.24±0.72          & 81.20±0.44          & 92.03±1.49          & 93.78±0.38          & 93.70±0.56          & 6                  \\ \cline{3-9} 
		& \textbf{RDD}     & 83.68±0.40          & 73.64±0.50          & 81.74±0.44          & 92.18±1.45          & 94.20±0.48          & 94.14±0.39          & 3.5                \\ \cline{3-9} 
		& \textbf{FreeKD}  & 83.84±0.47          & 73.92±0.47          & 81.48±0.38          & 92.38±1.54          & 93.65±0.43          & 93.87±0.48          & 4                  \\ \hline
		\multirow{6}{*}{\textbf{GNN-to-MLP}}   & \textbf{GLNN}    & 82.20±0.73          & 71.72±0.30          & 80.16±0.20          & 91.42±1.61          & 92.22±0.72          & 93.11±0.39          & 10.5               \\ \cline{3-9} 
		& \textbf{CPF}     & 83.56±0.48          & 72.98±0.60          & 81.54±0.47          & 91.70±1.50          & 93.42±0.48          & 93.47±0.41          & 6.83               \\ \cline{3-9} 
		& \textbf{RKD-MLP} & 82.68±0.45          & 73.42±0.45          & 81.32±0.32          & 91.28±1.48          & 93.16±0.64          & 93.26±0.37          & 8.17               \\ \cline{3-9} 
		& \textbf{FF-G2M}  & 84.06±0.43          & 73.85±0.51          & 81.62±0.37          & 91.84±1.42          & 93.35±0.55          & 93.59±0.43          & 5                  \\ \cline{3-9} 
		& \textbf{KRD}     & 84.1±0.87           & 75.26±0.37          & 82.14±0.52          & 92.22±2.14          & 93.86±0.38          & 94.70±0.37          & 2.33               \\ \cline{3-9} 
		& \textbf{SA-DSD}  & \textbf{85.22±0.66} & \textbf{75.33±1.18} & \textbf{82.81±0.47} & \textbf{93.48±1.45} & \textbf{94.11±0.46} & \textbf{95.31±0.29} & \textbf{1.17}      \\ \hline\hline
	\end{tabular}
\end{table*}

\subsection{Ablation Study}
\subsubsection{Evaluation of the Distillation Strategy}
\begin{figure*}[hbtp!]
	\centering
	\includegraphics[width=1\linewidth]{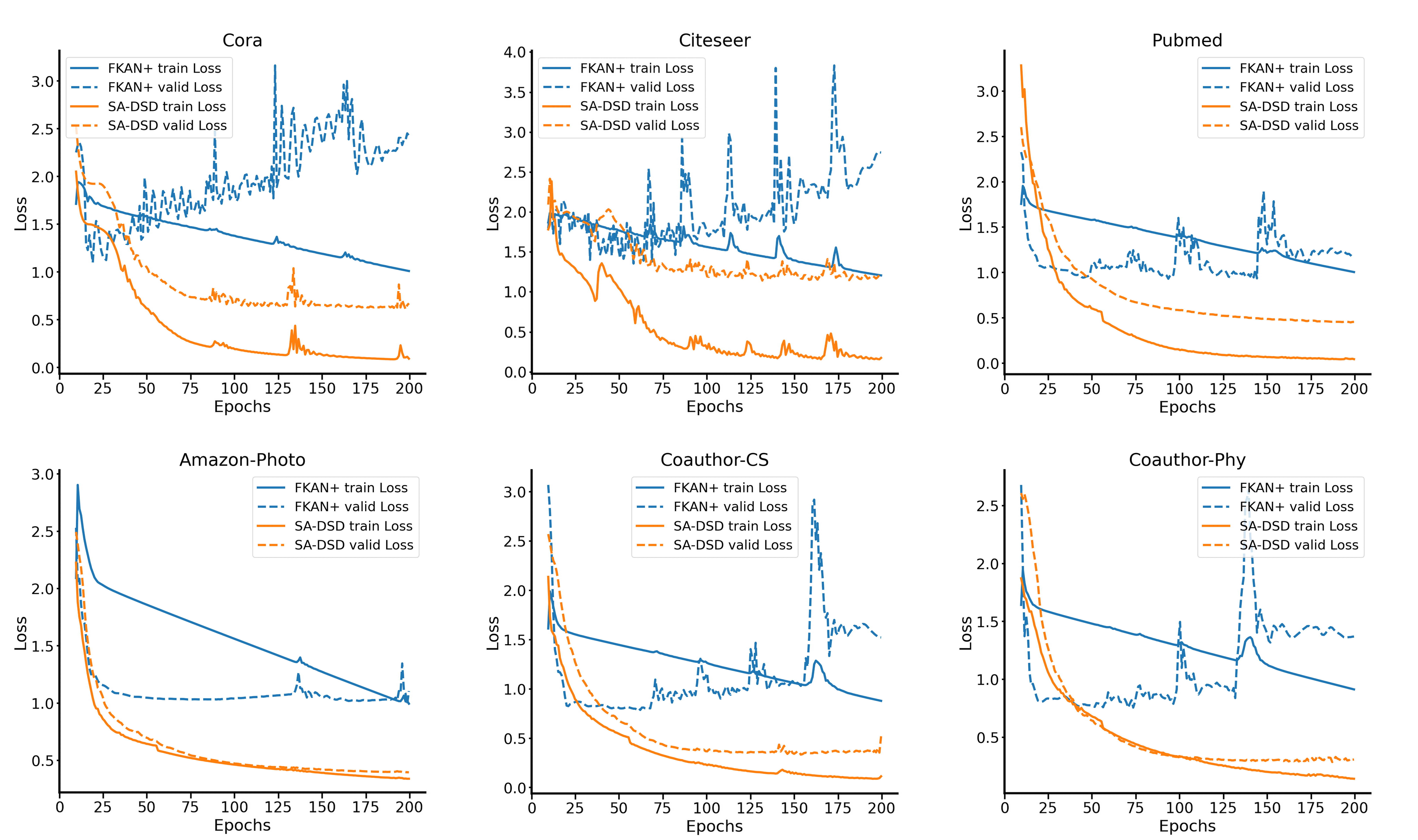}
	\caption{The loss curves of SA-DSD and FR-KAN+ on six datasets.}
	\label{fig:loss}
\end{figure*}

We found that the predictive performance of FR-KAN+ was significantly lower than that of GNN models with graph aggregation capabilities. However, by introducing the knowledge distillation strategy within the SA-DSD framework, its performance exceeded that of the GNN models. We further explored the reasons behind this discrepancy. As shown in Fig.\ref{fig:loss}, there is a noticeable gap between the training and validation loss curves of FR-KAN+, indicating a severe overfitting issue. In contrast, SA-DSD exhibited better convergence, particularly as the graph structure complexity increased, with a significant reduction in the fluctuation of training and validation losses. This result suggests that SA-DSD effectively models the nonlinear relationship between feature inputs and prediction outputs, with the distillation strategy playing a key role in suppressing overfitting and improving generalization. 

\begin{figure*}[hbtp!]
	\centering
	\includegraphics[width=1\linewidth]{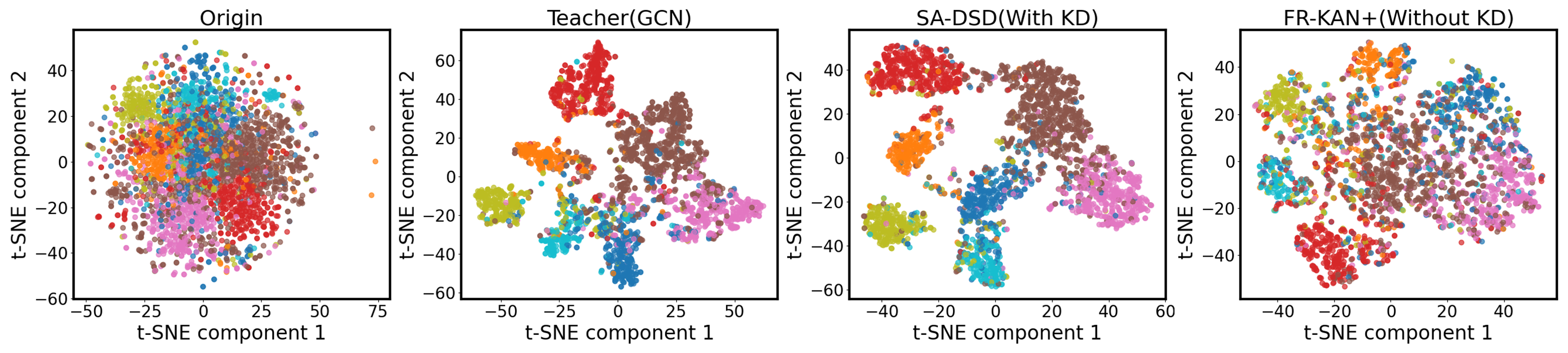}
	\caption{Visualization of model classification results.}
	\label{fig:t_sne}
\end{figure*}

Through t-SNE dimensionality reduction visualization in Fig.\ref{fig:t_sne}, we observed that the FR-KAN+ model without distillation performed poorly in category differentiation, with noticeable overlap between category clusters. On the other hand, in the feature embedding space of SA-DSD, the 7 categories were clearly separated, with greater separation than that achieved by the GNN teacher model. In other words, the distillation strategy can effectively compensate for the lack of neighborhood aggregation in FR-KAN+.

\subsubsection{Evaluation of the Student Model Improvement Gains}
\begin{figure}[htbp!]
	\centering
	\includegraphics[width=0.5\linewidth]{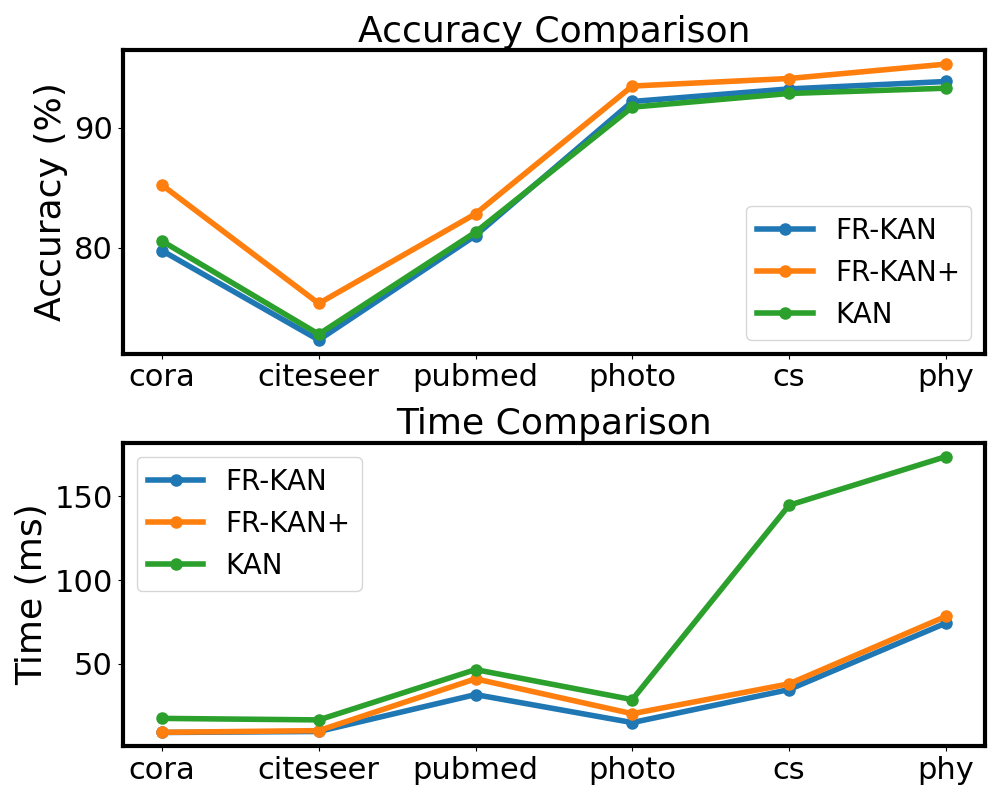}
	\caption{Comparison of the performance of the SA-DSD framework with different student models in the six datasets.}
	\label{fig:student_comparison}
\end{figure}

To verify whether the improved FR-KAN+ leads to performance gains, we conducted experiments using traditional KAN, FR-KAN, and FR-KAN+ as student models within the SA-DSD distillation framework. The experimental results are presented in Fig.\ref{fig:student_comparison}. The results show that FR-KAN significantly reduces computation time compared to traditional KAN, especially on datasets with large graph structures, where the time savings are more pronounced. However, the accuracy difference between the two models is minimal. With a slight increase in computational burden, FR-KAN+ significantly improves prediction accuracy. This indicates that the improvement of FR-KAN significantly enhances the performance of the model within the SA-DSD distillation framework, and the improvement is effective.

\subsubsection{Hyperparameter Sensitivity Analysis with Fixed Student Model}

\begin{figure}[htbp!]
	\centering
	\includegraphics[width=0.8\linewidth]{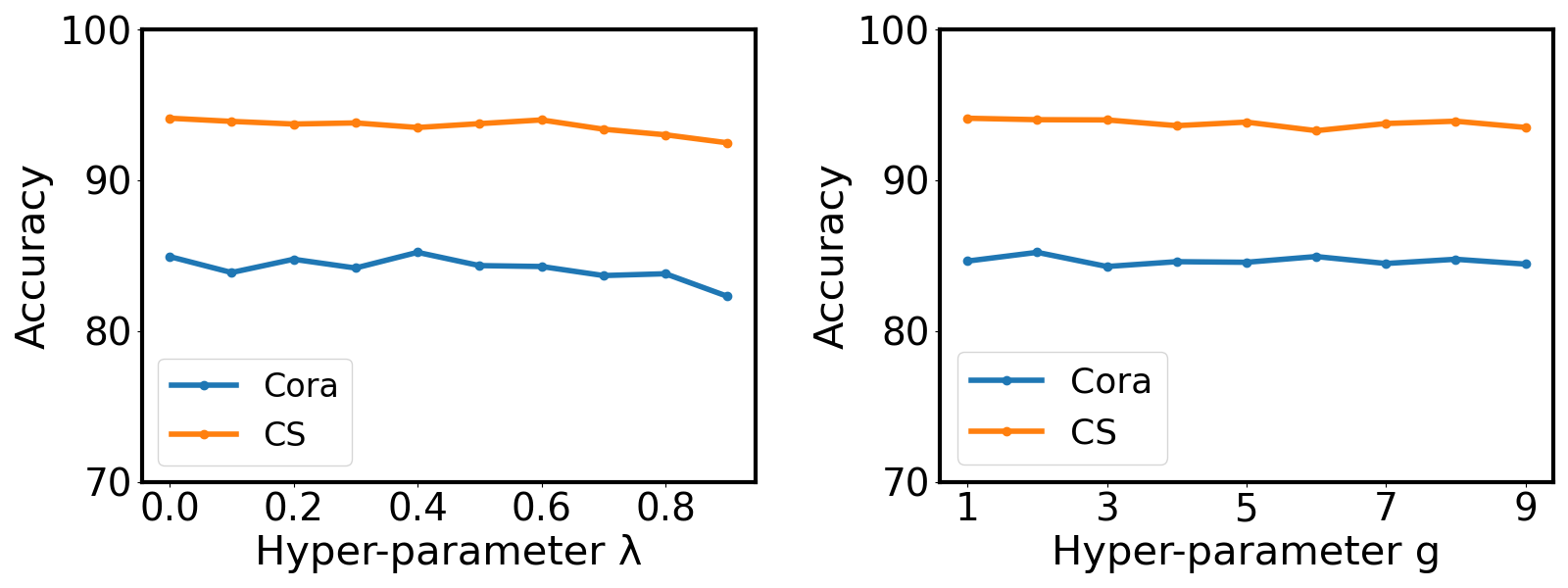}
	\caption{Hyper-parameter sensitivity analysis on \( \lambda \) and \( g \) under the SA-DSD framework, where the student model is fixed to FR-KAN+ and other settings follow the default configuration used in Fig.~\ref{fig:student_comparison}.}
	
	\label{fig:Hyper_parameter_sensitivity}
\end{figure}

We further analyze the sensitivity of the framework to \( \lambda \) and \( g \) on Cora and CS with GCN as the teacher and FR-KAN+ as the student. As shown in Fig.\ref{fig:Hyper_parameter_sensitivity}, we observed that when the value of \( \lambda \) is too large, model performance decreases. Increasing the value of \( g \), the number of Fourier bases, does not lead to significant performance improvements; rather, it substantially increases the model's parameter count. In practical applications, setting \( g \) to 1 yields good performance, while \( \lambda \) should be adaptively adjusted within a reasonable range based on training dynamics to ensure the model's generalization ability.

\section{CONCLUSION}
\label{CONCLUSION}
In this paper, we explore the use of edge-activated KANs as a replacement for fully connected MLPs in the context of knowledge distillation, and demonstrate its application potential in edge scenarios. We introduce a novel Self-Attention Dynamic Sampling Distillation (SA-DSD) framework. To the best of our knowledge, this represents the first attempt to employ GNN-to-KAN knowledge distillation. Specifically, we propose the FR-KAN+ model, which extends the traditional FR-KAN framework by integrating complex weights and Fourier transforms with dynamically adjustable frequency and phase shifts. This integration enhances computational efficiency while facilitating more effective frequency domain feature extraction. Additionally, we incorporate a self-attention mechanism to dynamically compute node level importance and reweight distillation signals throughout the distillation process. By applying upsampling based on the consistency between the predictions of the teacher and student models, we ensure robust knowledge transfer. This method effectively mitigates the aggregation limitations of the FR-KAN+ model. Extensive experiments and ablation studies across six real world datasets validate the efficacy of the proposed method and architecture. These results demonstrate the feasibility of distilling GNN knowledge into KAN style students under architectural mismatch.

Future work will explore more scalable attention designs for large graphs,
extensions to heterogeneous or dynamic graphs, and deeper analysis of when and why KAN-based students benefit most from graph distillation.


\section*{Funding}
This work was supported in part by the National Natural Science Foundation of China (U2133205).

\section*{Declaration of competing interest}
The authors declare that they have no known competing financial interests or personal relationships that could have appeared to influence the work reported in this paper.

\section*{Data availability}
Data will be made available on request.

\bibliographystyle{elsarticle-num} 
\bibliography{sa-dsd}

\appendix
\section{Hyperparameter Configurations}
\label{app:hyperparams}

To ensure reproducibility, we report the best hyperparameters of SA-DSD with FR-KAN+ as the student model obtained via Optuna optimization. These configurations are specific to each teacher model (GCN, GraphSAGE, GAT), dataset, and learning mode (transductive/inductive). The key hyperparameters include the distillation weight number of layers, hidden dimension, learning rate (lr), and grid size.

\begin{table}[htbp]
	\centering
	\footnotesize
	\caption{Best hyperparameter configurations of SA-DSD (Optuna).}
	\label{tab:hyperparams_app}
	\begin{tabular}{>{\centering\arraybackslash}m{1.5cm}
			>{\centering\arraybackslash}m{1.5cm}
			|*{6}{>{\centering\arraybackslash}m{1.8cm}}}
		\hline\hline
		Teacher & Mode & Cora & Citeseer & PubMed & Photo & CS & Physics \\
		\hline
		\multirow{2}{*}{\centering GCN}
		& {\centering Trans} & {\centering lay=2 \\ hid=32 \\ lr=0.005 \\ g=2}
		& {\centering lay=2 \\ hid=128 \\ lr=0.005 \\ g=10 }
		& {\centering lay=2 \\ hid=128 \\ lr=0.005 \\ g=5 }
		& {\centering lay=3 \\ hid=64 \\ lr=0.001 \\ g=1 }
		& {\centering lay=2 \\ hid=128 \\ lr=0.0005 \\ g=5 }
		& {\centering lay=3 \\ hid=128 \\ lr=0.001 \\ g=5 }\\
		\cline{2-8}
		& {\centering Ind} & {\centering lay=3 \\ hid=128 \\ lr=0.001 \\ g=10 }
		& {\centering lay=2 \\ hid=64 \\ lr=0.005 \\ g=6 }
		& {\centering lay=2 \\ hid=64 \\ lr=0.005 \\ g=6 }
		& {\centering lay=2 \\ hid=128 \\ lr=0.001 \\ g=1 }
		& {\centering lay=2 \\ hid=128 \\ lr=0.0005 \\ g=1 }
		& {\centering lay=2 \\ hid=64 \\ lr=0.001 \\ g=1 } \\
		\hline
		\multirow{2}{*}{\centering SAGE}
		& {\centering Trans} & {\centering hid=64 \\ lr=0.005 \\ g=3 }
		& {\centering lay=3 \\ hid=128 \\ lr=0.01 \\ g=3 }
		& {\centering lay=3 \\ hid=64 \\ lr=0.01 \\ g=9 }
		& {\centering lay=2 \\ hid=128 \\ lr=0.001 \\ g=8 }
		& {\centering lay=2 \\ hid=64 \\ lr=0.0005 \\ g=4 }
		& {\centering lay=2 \\ hid=128 \\ lr=0.005 \\ g=5 } \\
		\cline{2-8}
		& {\centering Ind} & {\centering lay=2 \\ hid=64 \\ lr=0.005 \\ g=6 }
		& {\centering lay=2 \\ hid=128 \\ lr=0.005 \\ g=10 }
		& {\centering lay=3 \\ hid=128 \\ lr=0.05 \\ g=4 }
		& {\centering lay=3 \\ hid=64 \\ lr=0.005 \\ g=1 }
		& {\centering lay=2 \\ hid=128 \\ lr=0.0005 \\ g=1 }
		& {\centering lay=2 \\ hid=128 \\ lr=0.0005 \\ g=4 } \\
		\hline
		\multirow{2}{*}{\centering GAT}
		& {\centering Trans} & {\centering lay=2 \\ hid=32 \\ lr=0.005 \\ g=8 }
		& {\centering lay=2 \\ hid=64 \\ lr=0.01 \\ g=4 }
		& {\centering lay=3 \\ hid=128 \\ lr=0.005 \\ g=10 }
		& {\centering lay=3 \\ hid=128 \\ lr=0.01 \\ g=1 }
		& {\centering lay=2 \\ hid=64 \\ lr=0.0005 \\ g=1 }
		& {\centering lay=2 \\ hid=64 \\ lr=0.001 \\ g=3 } \\
		\cline{2-8}
		& {\centering Ind} & {\centering lay=2 \\ hid=128 \\ lr=0.005 \\ g=10 }
		& {\centering lay=3 \\ hid=64 \\ lr=0.01 \\ g=3 }
		& {\centering lay=3 \\ hid=64 \\ lr=0.005 \\ g=1 }
		& {\centering lay=2 \\ hid=64 \\ lr=0.05 \\ g=1 }
		& {\centering lay=2 \\ hid=128 \\ lr=0.0005 \\ g=1 }
		& {\centering lay=2 \\ hid=64 \\ lr=0.001 \\ g=4 } \\
		\hline\hline
	\end{tabular}
\end{table}

\end{document}